\documentclass[conference]{IEEEtran}
\IEEEoverridecommandlockouts
\usepackage{cite}
\usepackage{amsmath,amssymb,amsfonts}
\usepackage{algorithmic}
\usepackage{graphicx}
\usepackage{textcomp}
\usepackage{xcolor}
\def\BibTeX{{\rm B\kern-.05em{\sc i\kern-.025em b}\kern-.08em
    T\kern-.1667em\lower.7ex\hbox{E}\kern-.125emX}}
\begin{document}

\title{A Survey on Visual Map Localization Using LiDARs and Cameras}

\author{Mahdi Elhousni,~\IEEEmembership{Student Member,~IEEE,}
        Xinming Huang,~\IEEEmembership{Senior Member,~IEEE,}
\IEEEcompsocitemizethanks{\IEEEcompsocthanksitem Both authors are with the Electrical and Computer Engineering department at Worcester Polytechnic Institute in Worcester, MA, USA.}}

\maketitle

\begin{abstract}
As the autonomous driving industry is slowly maturing, visual map localization is quickly becoming the standard approach to localize cars as accurately as possible. Owing to the rich data returned by visual sensors such as cameras or LiDARs, researchers are able to build different types of maps with various levels of details, and use them to achieve high levels of vehicle localization accuracy and stability in urban environments. Contrary to the popular SLAM approaches, visual map localization relies on pre-built maps, and is focused solely on improving the localization accuracy by avoiding error accumulation or drift. We define visual map localization as a two-stage process. At the stage of place recognition, the initial position of the vehicle in the map is determined by comparing the visual sensor output with a set of geo-tagged map regions of interest. Subsequently, at the stage of  map metric localization, the vehicle is tracked while it moves across the map by continuously aligning the visual sensors' output with the current area of the map that is being traversed. In this paper, we survey, discuss and compare the latest methods for LiDAR based, camera based and cross-modal visual map localization for both stages, in an effort to highlight the strength and weakness of each approach.
\end{abstract}


\begin{figure*}
        \begin{center}
        \includegraphics[width=0.9\linewidth]{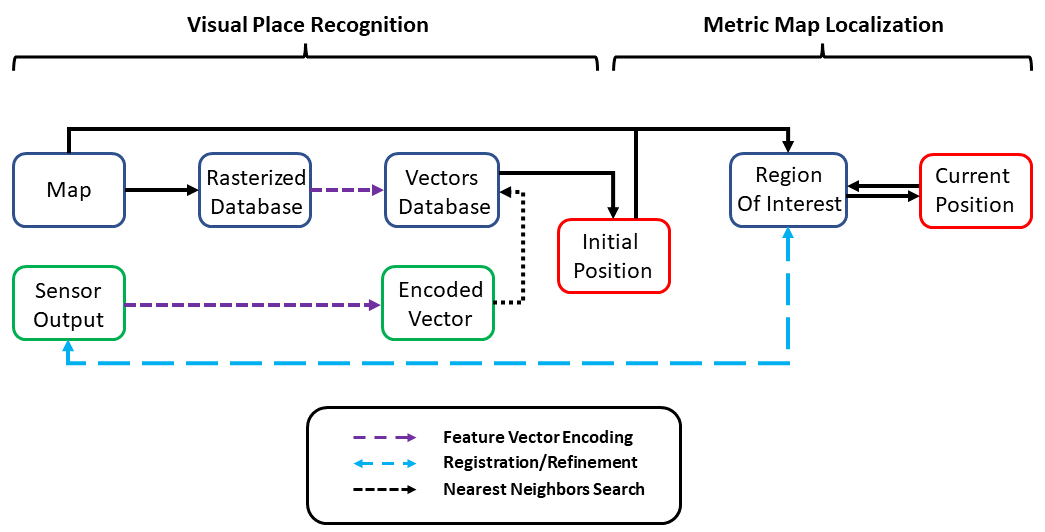}
        \caption{Visual Map Localization Block Diagram: First, the Visual Place Recognition stage where the map is rasterized to produce a database of geo-tagged samples for more efficient processing. This is followed by the encoding into feature vectors of both the sensor output and all the samples from the map. A nearest neighbor search is then used to find the closest map sample to the sensor output, and thus produce a guess at the initial position of the vehicle in the map. Next is the Metric Map Localization stage where a registration algorithm is used to align the sensor output with the map, making it possible to track the vehicle. }
    \end{center}
\end{figure*}

\section{Introduction}
Localization has become one of the corner stones of any modern robotics system, especially in the context of autonomous driving. Accurately localizing a vehicle can have a dramatic effect on subsequent tasks such as behavioral planning or moving object tracking. \par 
While it is possible to achieve reasonable results with traditional dynamic sensors such as IMUs and GPS,  modern robotics systems have shown that visual based sensors, mainly laser or camera based, are extremely well suited for this task, since the localization results returned by such sensors do not only depend on the robot itself, but also on its surrounding environment: Laser based sensors such as LiDAR are capable of providing accurate metric measurements to all the objects present on the line of sight of the sensors, making it uniquely adapted to the mapping aspect of autonomous driving, and as a consequence, capable of accurately achieving the localization task as well. On the other hand, camera-based sensors are capable of capturing rich texture-based keypoints, which can be matched across frames, and used as reference to calculate the displacement of the robot equipped with it. 
While both types of sensors possess numerous qualities that result into a good localization performance, they also suffer from some limitations. For instance, for the laser-based sensors, the inability to capture colors and textures may sometimes introduce ambiguity during the point matching process, and for the camera-based sensors, sudden changes in the brightness levels or the absence of an adequate lighting source can cause complete failure of the localization system. Such limitation can sometimes be overcame by using sensor fusion or cross-modal approaches. \par 
With a rich background in the robotics literature, localization has been explored through the years, most of time jointly with the mapping aspect, giving eventually birth to two distinct schools of thoughts: The first one is traditionally called Simultaneous Localization and Mapping, or SLAM where both the localization and mapping are executed simultaneously in a loop, making it possible for each one to take advantage of the results of the other. The second approach decouples the localization and mapping, by doing the latter offline, and using those results to achieve accurate localization. While SLAM can be seen as the most optimal solution in terms of deployment time, the accuracy attained by such systems is simply not enough to satisfy the safety conditions that would be necessary in order to deploy autonomous driving cars in urban environments due to the drift from which they sometimes suffer. In addition to that, obstacle detection and tracking methods are also still striving to achieve consistent results that could be trusted, even more when problems such as occlusion or sudden brightness changes arise, making it usually necessary to use the pre-built maps as a platform for labeling and eventually detecting relevant environment information, such as traffic lights and traffic signs. \par
Visual map localization can typically be divided into two major steps. First, the vehicle (or robot) must find its initial location on the map, especially when no other sensor such as a GPS is available to provide an initial guess or a region of interest. The solution in this case is to use the Visual Place Recognition approach, where using only the input of our visual sensor and an intermediate representation, we can find the best match in the pre-built map. Once the initial location is found, the robot can now start to navigate the map, while we track its movements as accurately as possible. We call this step Map Metric Localization which is achieved by enforcing both a temporal consistency between the subsequent frames provided by the input sensor, in addition to a spatial consistency, which is guaranteed by matching with the map’s region of interest and can be seen as a correction to the first transformation that was calculates using the sensors inputs only. This second step runs in a recurrent fashion, as long as the localization error stays at a reasonable level, guaranteeing enough overlap between the sensor’s outputs and the map’s region of interest. Fig. 1 shows a block diagram representing the different processes involved in each of the two major steps discussed above.\par
In this survey, we will explore all the major visual place recognition and metric map localization methods available in the literature, for both laser and camera-based sensors. The results provided by each publication will be presented and compared, using established benchmarks, to try and find the best solutions for this essential task, and discuss the strengths of weaknesses of each one. \par

\textbf{Scope:} This survey paper focuses exclusively on localization methods relying on offline pre-built maps using LiDARs or cameras as input sensors, which excludes SLAM (Simultaneous Localization And Mapping) approaches. We also consider the case of ``cross-modal" based localization but make a clear distinction with ``sensor fusion" based methods: cross-modal methods use one sensor output and attempts to localize it in a map that was constructed using a different modality sensor. On the other hand, sensor fusion methods use the inputs of sensors from both modalities, regardless of the map type. Sensor fusion methods and map-less localization methods were already complied in \cite{debeunne2020review,elhousni2020survey,wu2018image} and hereby will not be discussed in this survey, where we focus only on single-sensor, single-map localization methods.\par
\textbf{Outline:} This paper will be structured as follows. In Section II, we will define the two major steps in visual map localization, namely place recognition and metric map localization. These two steps will then be presented in detail in Section III and IV, by considering for the following sensor configurations: camera based, LiDAR based and cross-modal based methods. Finally in Section V, we will present the results on the discussed methods and compare them using multiple existing datasets.  \par

\section{Background}
In this section, we will discuss the two main steps in most visual map localization approaches: Place recognition, to find the initial position in the map, then the map metric localization, to keep track of the vehicle while it moves in the map.

\begin{table}[h]

\begin{centering}


  \caption{Visual map examples.}

\begin{tabular}{|*{2}{c|}}
      \hline
      {LiDAR Maps} & {Camera Maps}\\
      \hline
      {Intensity Maps} & {Satellite Maps}\\
      {Pointcloud Maps} & {OpenStreetMaps}\\
      {Mesh Maps} & {Google StreetView}\\
      {HD Maps} & {Depth Maps}\\
      \hline

  \end{tabular}

\end{centering}
\end{table}

\begin{figure*}
        \begin{center}
        \includegraphics[width=\linewidth]{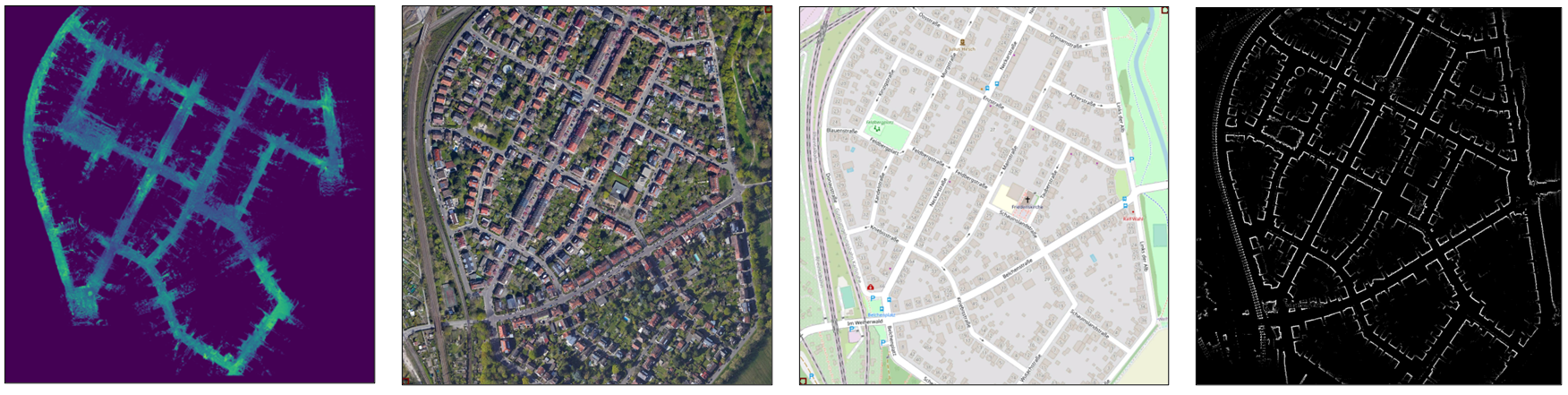}
        \caption{Visual Maps. From left to right: LiDAR map, satellite map, OSM and simulated LiDAR map.}
    \end{center}
\end{figure*}

\subsection{Visual Place Recognition}
Visual place recognition represents the task of finding the best match possible for a visual input sensor, in a pre-built database, as efficiently as possible. This task has multiple applications such as image recommendation for search engines or loop closure for SLAM systems. In our case, we are interested in the methods which use visual place recognition as a means to find the initial position of a robot in a pre-built map. \par 
Tab. 1 lists a few examples of the visual maps available nowadays. Fig. 2 shows a set of pre-built maps covering the same areas (Sequence 00 of the KITTI Odometry Dataset \cite{geiger2013vision}) but constructed using different methods. First, the LiDAR map, built using successive LiDAR point clouds which were aligned using registration methods, and fine-tuned using bundle adjustment online. Next is a satellite map, built using satellite imagery and GPS measurements, followed by OpenStreetMap (OSM), which are constructed by tracing relevant objects over satellite maps. Finally, a simulated 2D LiDAR map, constructed by applying raycasting to a buildings-segmented-OSM. \par
Visual Place Recognition is typically approached differently, depending on the sensors that is being used: When cameras are used to find the initial position, the map used is typically a variation of a top view 2D map, such as satellite-based maps. While these maps tend to have limited sizes, and thus are easily rasterized and queried, they do introduce another challenge which is the cross-view aspect: maps are captured from the top view, while the camera is typically mounted to capture either a front view, or a panoramic 360 view, but always from the ground level. This challenge is usually solved using deep learning techniques, which focus more on textures, rather than the physical aspect of the structures present in the image. On the other hand, laser sensors do not suffer from any view discrepancy, since it is much easier to project their output to a new view. This makes it possible to use both traditionally handcrafted and deep learning methods. However, when dealing with laser-based sensors, speed can sometimes be an issue, because processing all the dense data from the map into an easily searchable database can take more time. \par
Fig. 3 and Fig. 4 show the different projections that could be used to represent the output of the visual sensors discussed in this review. For LiDARs, Fig. 3 shows the typical point cloud view, the BEV projection view and finally the panoramic projection view. For the camera, Fig. 4 shows the front-view camera, the satellite view, the OSM view and finally the so-called polar projection of the satellite view, which aims to look as close as possible to a ground view.

\subsection{Metric Map Localization}

Once the initial position in the map is found, the vehicle must now keep track of its current position on the map. This, depending on the map and the sensor used, can be done following different approaches. In the case of LiDAR localization in point cloud maps, localization is typically achieved using some form of point cloud registration, which represents the task of aligning two point clouds by finding the 3D spatial transformation between both scans. Accurate keypoints extraction and matching tend to be essential to obtain an accurate transformation, however, newer end-to-end deep learning-based methods such as \cite{chen2021overlapnet} claim to be able to bypass that. On the other hand, camera-based map localization for autonomous driving cars tends to be more challenging, once again due to the drastic change in viewpoint. Traditional computer vision techniques for keypoint extraction and matching typically fail when attempting to locate ground captured images in aerial maps, which has pushed researchers to rely more heavily on deep learning to solve this challenge. \par
Finally, it is possible to combine both sensors to solve the metric map localization challenge: Predicting a depth map using monocular camera images or using raycasting to produce simulated point clouds are some of the usual methods used to break the cross-modality issue between LiDARs and cameras.

\section{Visual Place Recognition}

In this section we will discuss the methods that attempted to solve the typical first stage of most visual map localization methods where the initial position of the vehicle on the map is determined: visual place recognition. Here, three sensor configurations will be considered: LiDARs, cameras and cross-modal based approaches.

\subsection{LiDAR Based Methods}

LiDAR place recognition has become very popular since HD point cloud maps have become the norm for many autonomous driving vehicles. The earlier attempts to solve this task tried to capitalize on the advances in keypoint detection and matching for point clouds. In \cite{bosse2013place}, based on a random sampling procedure, keypoints were selected and encoded using a variation of the gestalt
descriptor \cite{bileschi2007image}, before being matched using the nearest neighbor voting approach. In \cite{himstedt2014large}, the keypoints based place recognition task was solved by taking advantage of the geometrical relations between points: after extracting features using points of high curvature, the authors encoded the point cloud data into a 2D histogram based on the distances between them and their co-bearings, which resulted in a signature that was later used to match the point cloud with other scenes using the Approximate Nearest Neighbor Search. In \cite{dube2017segmatch} the authors proposed SegMatch, an algorithm based on segmentation results which were then used to construct feature descriptors. The matching of segments was achieved following a two-step approach: first using a random forest classifier, followed by a geometrical verification using RANSAC \cite{fischler1981random}. This was eventually extended in \cite{vidanapathirana2021locus}, by augmenting the SegMatch descriptor with a handcrafted spatiotemporal descriptor which was constructed following two stages of spatial and temporal feature pooling.  \par

\begin{figure}
        \begin{center}
        \includegraphics[width=\linewidth]{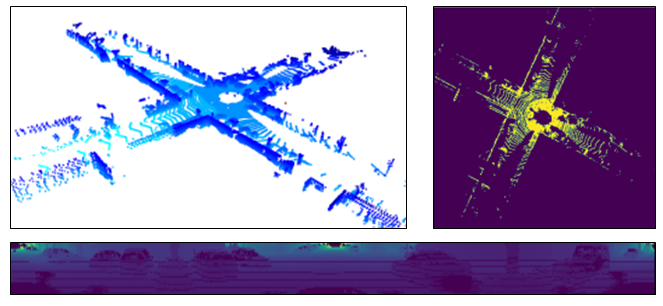}
        \caption{An example of LiDAR frame depicted as a 3D point cloud (top left), a BEV projection (top right) and a panoramic projection (bottom).}
    \end{center}
\end{figure}

The authors of \cite{guo2019local} proposed to take advantage of the intensity field returned by the LiDAR sensor to construct an intensity-augmented 3D keypoint descriptor named ISHOT, which was matched following a strategy combining probabilistic voting and nearest neighbor search. A similar method was used in \cite{shan2021robust} where the intensity field was central to the approach, but in this work, the intensity data was first projected to the 2D image space using a panoramic projection, before using a traditional computer vision (CV) keypoint extractor and encoder, in this case ORB \cite{ rublee2011orb}. This was followed by a traditional CV matching procedure relying on PnP \cite{ fischler1981random} and BoW \cite{ sivic2008efficient}. Projecting 3D point clouds to 2D in order to take advantage of traditional CV techniques is a common method used when processing LiDAR data. Another method that utilizes this principle was proposed in \cite{luo2021bvmatch}, but this time, the Bird Eye View (BEV) projection was used. An appropriate descriptor named BVFT was proposed, and similarly to the previous discussed method, a BoW matching method was deployed, followed by ICP \cite{Zhang94iterativepoint} refinement. Lately, approaches relying on encoding the full scan into some sort of compressed representation have become more popular, which resulted in the development of the popular ScanContext \cite{kim2018scan} encoder. In this work, the authors proposed a two-step process which results in a compressed and viewpoint invariant 3D tensor, where the position, orientation and height of each point were encoded. The resulting global descriptors were matched using a simple similarity score. \par
Lately, deep learning has been increasingly used to try and solve the LiDAR place recognition task, first by including it into semi-handcrafted methods such as \cite{yin2018locnet,yin20193d}, where the point clouds were first pre-processed using a histogram based method to produce rotation invariant representations, which were then fed to a siamese neural networks, trained using the contrastive loss function in order to generate similar vector representations for similar point clouds. Likewise, the authors in \cite{li2022rinet} followed a similar strategy by first generating a rotation invariant representation, based on the semantic segmentation of the overhead projection of the point clouds, followed by a siamese neural network for feature extraction, and a MLP for similarity prediction. Another semi-handcrafted method was proposed in \cite{li2021lidar}. Here the authors started by generating an overhead projection of the point clouds, then processed them in order to generate two types of descriptors: a global one, generated using the NETVlad architecture \cite{arandjelovic2016netvlad}, and a feature based one, generated using the SuperPoint \cite{detone2018superpoint} architecture. Both descriptors were combined, and matching was achieved using the SuperGlue algorithm \cite{sarlin2020superglue}. End-to-end methods attempting to solve this problem have been proposed too, notably in \cite{fischer2022stickylocalization}, based on the combination of a graph neural layer with an optimal transport layer. The network was then trained using a distance-based matching loss that rewards
closer points and penalize farther ones, instead of the typical binary ground truth used for matching. Graph neural networks were also used in \cite{kong2020semantic}. Here the graph was generated based on semantic segmentation results of the point clouds, then fed into a graph neural network with the following steps: node embedding, graph embedding and graph-graph interaction.

\subsection{Camera Based Methods}

As mentioned before, camera-based place recognition (also called Camera Cross-View Localization) can be very challenging due to the large difference in viewpoint between the images collected by the ground vehicles, and the images extracted from the aerial maps. Consequently, most of the popular and successful methods rely on deep learning. This was first demonstrated in \cite{tian2017cross}, by relying on a Faster R-CNN \cite{ren2015faster} to detect buildings then match them using a siamese network trained using the contrastive loss. Note that both the contrastive and triplet losses are very popular when trying to solve this challenge as we will see in the following cited publications. This was improved upon the following year in \cite{hu2018cvm} by simplifying the first stage from object detection to CNN feature extraction followed by an encoding stage using the NetVLAD architecture \cite{arandjelovic2016netvlad}. The two previously cited works established a common basis which was typically used as a starting point to the methods that followed. \par 

\begin{figure}
        \begin{center}
        \includegraphics[width=0.90\linewidth]{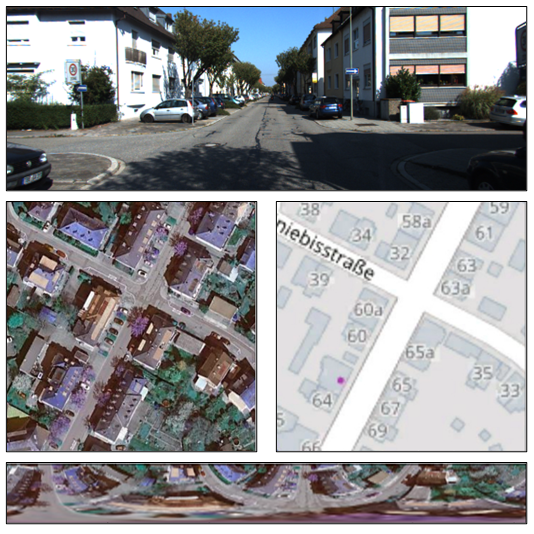}
        \caption{Front view camera frame (top), followed by the same area in a satellite map (middle left) and in OSM (middle right). The final image is the polar projection of the satellite crop (bottom).}
    \end{center}
\end{figure}

In \cite{liu2019lending}, the authors proposed to attach a color encoded orientation map to the input queries during training and testing, which seems to improve the accuracy on the most challenging metrics. The importance of orientation alignment was furthermore represented in \cite{zhu2021revisiting} where the authors showed that training using images that were pre-aligned first in terms of orientation will produce a siamese network that is capable of producing activation maps that perform better at pointing similar objects in different views. The activation maps, which were produced using GRAD-CAM \cite{selvaraju2017grad} can also be used during testing to approximate the orientation that best aligns the two views. Another approach to improve the accuracy of a siamese network trained for cross-view geo-localization is to take advantage of the results of traditional semantic segmentation networks and include them in the data augmentation procedure during training: this was done by removing different segmented objects in the ground images, as a way to make the network more robust to temporal changes in the images. This, combined with a multi-scale attention module, produces better ranking and matching results. \par 
The authors of \cite{shi2020optimal} introduced the use of optimal features transport \cite{cuturi2013sinkhorn} to facilitate the extraction of similar features in both views. This was implemented in a way that allowed the end-to-end training on the network and showed great improvements across all metrics. \par
While most works use some sort of variation of the contrastive or triplet losses, the authors in \cite{guo2022soft} proposed their own metric, dubbed Soft Exemplar Highlighting Loss. In their formulation, this loss, combined with a polar transform applied to the aerial images to reduce the viewpoint gap, was used to assign different weights to the training examples depending on their difficulty, in an effort to emphasize meaningful hard samples and remove problematic ones. Another typical assumption in most cross-view geo-localization works in the literature is the one-to-one matching assumption between aerial and ground images. This does not always hold during testing and was the main motivation in \cite{zhu2021vigor}: in this work, the authors did not only attempt to predict the matching score between two samples, but also using a regression branch, predicted a latitude and longitude-based offset between the two inputs. Also, in addition to the triplet and regression losses, the authors introduced an IOU-based loss to better learn from semi-positive sample (meaning aerial samples with a non-zero offset). \par 
Lately, because of the success of attention models in computer vision \cite{guo2022attention}, more works have been trying to use the attention mechanisms \cite{vaswani2017attention} and the Transformer architecture \cite{dosovitskiy2020image} to solve this task, starting with \cite{shi2019spatial} where the authors proposed to use what they call a Spatial-aware position embedding module to process both the ground and polar transformed aerial images, tasked with  encoding the relative positions among object features extracted by the backbone network. This module consists of a max pooling block, followed by two fully connected layers in order to select the most important features. In \cite{yang2021cross} the authors proposed an architecture where first, for both views, 1D learnable encodings were combined with a set a CNN extracted features, before being fed into what the authors called a Layer-To-Layer Transformer: basically, a transformer with skip connections between timesteps. In \cite{zhu2022transgeo}, the authors attempted a pure transformer architecture which does not make use of CNN's as pre-processing step for feature extraction: this was done by following a two-stage procedure, where in the first step, two traditional Vision Transformer (ViT) architectures were trained using the triplet loss to generate embedding features for both street and aerial views. In the second stage, the aerial attention map generated from the first stage was used as guidance to crop and zoom-in on the most relevant portion of the image. This new generated aerial image was then used to finetune the aerial embedding using another ViT.

\subsection{Cross-Modal Methods}

Because of the scarcity and lack of availability of accurate HD point clouds maps, researchers have been trying to solve the place recognition challenge when having a LiDAR point cloud as input by using freely available and sometimes opensource maps such as satellite maps or OpenStreetMaps (OSM). Solving this typically involves the use of deep learning since we not only have to deal with the gap in modality, but this is exacerbated by the gap in viewpoint too. Lately, the authors in \cite{tang2021get} proposed a method where based on a predicted occupancy map from a satellite image, raycasting was used to generate simulated overhead LiDAR images, which were then combined with the overhead projections of the sensor inputs and fed into a DGCNN architecture \cite{wang2019dynamic} to predict a transformation offset, but also in a NetVLAD architecture to generated embeddings that could be used for place recognition. In \cite{cho2022openstreetmap}, it was OSM that was used as main map. By taking advantage of the buildings and roads information’s, the authors used raycasting to generate simulated overhead LiDAR images, which were matched with the LiDAR sensor's input using the Scan-Context \cite{kim2018scan} descriptor discussed previously.

\section{Metric Map Localization}

In this section we will list the works that proposed to solve the most challenging stage of visual map localization: metric map localization. Using LiDARs, cameras and cross-modal based approaches, we will discuss how it is possible to track a vehicle traversing a map using a single visual sensor as accurately as possible.

\subsection{LiDAR Based Methods}
LiDAR localization using a pre-built map has been the most successful approach for autonomous driving vehicle in terms of accuracy. This is due to the rich amount of detail typically available in such maps, since every area is the result of multiple scans that were aligned and concatenated. 2D LiDAR localization has a long and rich amount of published research in the robotics community, especially for indoor scenarios. In contrast, we will mainly focus on 3D LiDAR which are more adapted to outdoor scenarios and are typically available in modern autonomous driving cars. \par
Earlier methods such as \cite{levinson2007map} relied on sensor fusion and particle filters to localize LiDAR equipped vehicles in point clouds maps. In \cite{yoneda2014lidar} a solution to LiDAR map localization was proposed through the design of handcrafted features that could be matched across the map and the sensor input point clouds and which were based on the histogram of the frequency of points clusters sizes. Some works such as \cite{castorena2017ground}, only relied on the intensity field returned by the LiDAR sensor, and in \cite{zhang2018robust} a method combining features and filters to deal with noisy LiDAR data due to rainy conditions was discussed: Feature extraction is based on the position and reflectivity of each point, followed by a combination of a particle filter (to process for vertical features) and a histogram filter (to process for ground features).  \par 
The authors in \cite{wolcott2015fast} drew inspiration from the NDT odometry algorithm \cite{magnusson2009three} and proposed to use Gaussian Mixture Maps (GMM). By using the ground plane $xy$ as a 2D grid, each cell in the grid can be filled using a one-dimensional Gaussian mixture that models the distribution over that cell’s height. An efficient multi-resolution branch-and-bound search was used to match cells and align the sensor point cloud with the map. Compressing the 3D map into a 2D representation to achieve faster results has also been explored in \cite{javanmardi2017autonomous} which proposed to use buildings footprints to generate a simplified segments-based map, which was then combined with NDT to solve the localization challenge. \par
If the authors are using the full 3D map, they sometimes have access to labels such as traffic lights or lanes, which can aid in the localization process. For example, the authors in \cite{ghallabi2018lidar} proposed to take advantage of the lane information to achieve lane-level accuracy using LiDARs. Roads were extracted mainly based on their height information, then lanes were detected using the intensity field returned by the LiDAR sensor. Finally, the map matching and pose tracking were achieved using a particle filter. \cite{ghallabi2019lidar} is an extension of the lane based localization but instead uses traffic signs (extracted using the points normals) as landmarks, and in \cite{chen2021pole}, authors used poles and curbs to localize the vehicle in a HD map. A pole cost function and a curb cost function were proposed and fused to generate a rough guess at the vehicle's position. \par

Deep learning is very popular when talking about place recognition, so naturally researchers try to use it with this task as well. First some method only rely on the results of other neural networks to improve their localization pipeline: In \cite{rozenberszki2020lol} a system that combines LiDAR odometry with segmap's place recognition to reduce the LiDAR position drift was proposed. This was achieved by taking advantage of the matched segments and aligning them in order to finetune the transformation obtained by the LiDAR odometry. In \cite{yue2020collaborative} the authors proposed a multi-vehicle collaborative approach aided by semantic segmentation. In the case of two vehicles for example, the proposed system enforces a geometrical and semantic consistency matching across the inputs of both vehicles. This produces a weighting matrix which subsequently used in an Expectation-Maximization algorithm to align the point clouds with the map. \par

End-to-end methods have also been proposed: the authors in \cite{chen2021overlapnet} used a siamese network, which processes the panoramic projection of different cues generated from the point clouds (semantic labels, point locations etc.) and predicts two quantities: a similarity score representing the overlap between both inputs and a relative yaw angle. The predictions were combined with a particle filter to achieve LiDAR map localization. In \cite{lu2019l3}, the authors proposed a network that attempts to learn the residual value between a traditional localization system and the ground truth. Relevant features were first extracted and fed into a miniPointNet \cite{qi2017pointnet} to generate their corresponding feature descriptors. A cost volume was then constructed in the solution space $(x, y, z)$ and regularized with 3D convolutional neural networks. Additionally, an RNN branch was added to the network structure to guarantee the temporal smoothness of the displacement’s predictions. Following the latest trends, \cite{engel2021attention} proposed to use attention mechanisms to solve the self-localization challenge in a point cloud HD map. The localization process was split in two phases: first, a landmarks association step where points association was achieved by combining kNN and local attention, followed by a global point cloud registration where the associations made in the first step were fed into a pose regression network which mainly contains a global attention/pooling layer followed by a MLP. \par

\subsection{Camera Based Methods}
When it comes to metric map localization using cameras, the task typically suffers from the same issues faced when attempting to first solve the place recognition step, meaning the drastic difference in viewpoint. In addition to that, we now also must deal with classical odometry and map localization challenges such as the accumulation of positional error or the lack of sufficient overlap between the map and the sensor output. One of the earliest solutions was proposed in \cite{brubaker2015map} based on the graph representation of the road network in OSM and the input of two cameras. Using the same setup, in \cite{ballardini2016leveraging}, the proposed approach relies on the buildings structure represented in OSM, rather than the road network. Here, buildings geometry was extracted from the input point clouds using filtering and clustering and scored against the OSM buildings data using a 2D scoring function based on orthogonality, in order to keep track of the vehicle position in OSM. \par 
The authors in \cite{wolcott2015fast} chose to use satellite maps instead. By using the depth information than can be generated using a stereo camera rig, the authors trained a Ground-Satellite Dictionary to be able match features from both views. Localization was achieved by first extracting features and their feature vectors from the ground views, then queering up the aerial images containing features with the closest feature vectors. In \cite{kim2017satellite}, only a single monocular camera was used to find the vehicle position in the satellite map. This was achieved by training a siamese neural network to predict a similarity score between ground images and aerial regions of interest (ROI). The predicted similarity score was then used to update the weights in a particle filter \cite{thrun2002probabilistic} in order to localize the monocular camera in the map. \par 
While multiple methods rely on extracting and matching visual features, others proposed to rely on extracting and matching visual landmarks instead. The landmarks used in \cite{spangenberg2016pole} were poles. The authors first started by constructing a pole map by detecting poles using the disparity image that can be generated using stereo cameras, combined with edge detection and logistic regression. Subsequently, localization was achieved by detecting poles in the same way, and then using that information to update a particle filter, which was coupled with a Kalman Filter \cite{thrun2002probabilistic} for additional sensor fusion. As an extension to \cite{shi2019spatial}, another sensor fusion method was proposed in \cite{xia2021cross} to take advantage of the noisy GPS measurements that are usually available: using a modified triplet loss function, the authors argue that the rough GPS measurements of the ground and polar transformed aerial images in a pre-defined region of interest could be used to calculate a weight capable of scaling the contribution of each pair of images accordingly. The effectiveness of the method was later demonstrated by combining it with a particle filter. The same authors proposed later a more advanced method in \cite{shi2022accurate} where in addition to the popular polar transform, they introduce a geometry-constrained projective transform that results in much more realistic ground looking images. In addition to that, a new fine-grained cross-matching solution was proposed: Based on the prediction of their baseline network, a corresponding aerial image was selected, tagged with a rough GPS location. The authors then proceed to transform the aerial image using their proposed projective transform and a set of pre-defined positions. Finally, the SSIM similarity loss function was used to select the best matching one. \par 
One final camera map representation, which is still sometimes used (although not very popular due to its sparsity), is Google StreetView. The authors of \cite{agarwal2015metric} transformed the closest panoramic image available in Google StreetView according to GPS to a set of eight rectilinear images, followed by a traditional homography-based feature matching, using SIFT features, to keep track of the vehicles position.
\subsection{Cross-Modal Methods}
Multi-modal approaches have been proposed to deal with certain edge cases such as autonomous driving cars that only have access to cameras but no LiDAR sensors, or for areas where we do not have HD point cloud maps available. This typically leads to the introduction and use of other pre-processing steps such as semantic segmentation, raycasting or depth prediction. \par
First, we will address methods that attempt to localize LiDAR point clouds on camera-based maps, such as satellite maps or OSM(-like) maps: In \cite{yan2019global}, the authors proposed a handcrafted 4-bit semantic descriptor, based on buildings and intersections positions in OSM cropped images and LiDAR semantic range images, which was combined with a particle filter to achieve global map localization. This work showed that semantic segmentation can be a great tool to break the multi-modality issue, and subsequent works took advantage of that, such as in \cite{s22145206} where the authors started by extracting the roads and buildings from both OSM and the input point clouds. These extracted regions were then used to generate BEV LiDAR point clouds images using the sensor inputs and BEV simulated point cloud images using OSM and raycasting. Finally, a road-constrained particle filter was used to align the different top view images and track the vehicle’s position in OSM. For satellite maps, the authors of \cite{miller2021any} also leverage the correlation of the semantic segmentation results from both the LiDAR point cloud and the satellite images in order to optimize the soft cost function of a particle filter. More advanced deep learning-based methods have recently been proposed: In \cite{tang2020rsl} and \cite{tang2020self}, a Generative Adversarial Network (GAN) \cite{creswell2018generative} was trained to generate synthetic top view LiDAR images based on input satellite crops. The synthetic and real LiDAR images were then both fed to a neural network to predict the value of the displacement between frames in a cascaded fashion, by first predicting the rotation value, then using that to predict the translation offset.\par
Next, we will discuss methods used to localize camera data in LiDAR maps. Stereo cameras are the natural pick when trying to localize video data in LiDAR maps because we can process their output to transform the data from 2D to 3D, which makes its alignment with point clouds much easier: Some of the early works attempting this include \cite{xu20173d} where using the point cloud maps, the authors started by generating synthetic depth images, then proceed to estimate the height, roll and pitch angle using the v-disparity map. This was followed by a dynamic object removal using the height information and finally the matching and alignment of the synthetic depth images with the stereo depth ones. Similarly, the method proposed in \cite{kim2018stereo} attempted to localize a stereo camera in a 3D LiDAR map, in this case by first relying on visual odometry to provide an initial guess at the transformation, before fine tuning it, using the synthetic and stereo depth maps residual alignment. \par
More challenging though is the task of localizing monocular camera images in 3D point cloud maps since they do not contain any depth or 3D information by definition. Some early attempts include \cite{caselitz2016monocular}, which showed that it is possible to use the results of bundle adjustment to generate a set of 3D points that can be referred to as a local reconstruction and later used to align the camera odometry with the 3D map following a similar scheme to ICP. Another method was proposed in \cite{wolcott2014visual} based on the idea of correlation between synthetic maps views and camera images. However, this time the synthetic images were populated using the intensity returned by the LiDAR sensor, instead of the depth data, which as a result produces synthetic images with a closer visual aspect to the camera images. Using a discreet number of possible synthetic images located around an initial pose guess, the authors used the Normalized Mutual Information (NMI) to evaluate them and determine the correct vehicle pose. Finding features which can be matched across both modalities can be challenging, however in \cite{yu2020monocular}, it was shown that features representing lines can help us achieve this objective. After detecting the ones in both 3D and 2D views, the authors excluded 3D lines that were not visible from the camera point of view, then proceeded to construct a feature vector for all the lines using various geometrical proprieties (such as length, orientation etc.) and matched them across both views. This makes it possible to track the camera images in the 3D maps. \par
As with all other challenges, solution involving deep learning were soon showing great potential: In \cite{cattaneo2019cmrnet} and \cite{cattaneo2020cmrnet++} the authors proposed CMRNET, a neural network capable of processing as input a RGB camera image and a synthetic depth map image and predicts as a result the relative pose between both inputs. A modified version of PWC-Net \cite{sun2018pwc} (an optical flow prediction network), was used, and the original method was later improved with the incorporation of PnP and RANSAC as a post-processing steps.

\section{Evaluation and Discussion}

Using popular datasets, we will proceed in this section to compare the major methods that were presented previously, for both stages of the visual map localization process, and using the same three modalities as we did before. Relevant conclusions will then be presented at the end, listing the strength and weaknesses of each modality configuration.

\subsection{Datasets} To compare the results reported by the previously cited methods, we selected the following datasets: 
\begin{itemize}
    \item {\em CVUSA \cite{workman2015wide}:} Consist of 44,416 pairs of panoramic ground-level and aerial/satellite images with normalized orientation and aligned GPS positions. The images were collected from across the United States and depict streets of both rural and urban scenes. This dataset in mainly used to evaluate place recognition methods.
    \item {\em CVACT \cite{liu2019lending}:} Similar to the CVUSA dataset, but contains 137,218 of pairs of images. This dataset too is mainly used to investigate place recognition approaches.
    \item {\em KITTI \cite{geiger2013vision}:} One of the most popular large scale dataset for outdoor odometry evaluation: It contains 22 sequences recorded using a Velodyne HDL-64E that was mounted on top of a car, resulting in LiDAR scans that were then pre-processed to compensate for the motion of the vehicle. Ground truth is available for the 11 first sequences and was obtained using an advanced GPS/INS system. This dataset can be used to evaluate both map metric localization methods and place recognition methods.
    \item {\em KITTI-360 \cite{liao2022kitti}:} An update to the largely successful KITTI dataset, recorded in the same city using a similar LiDAR setup and containing over 100k laser scans in a driving distance of 73.7km. Special attention to geo-localization alignment when building the dataset guarantees more accurate results when aligning with OSM or satellite maps. Similarly, to the previous KITTI dataset, this dataset too can be used to evaluate both map metric localization methods and place recognition methods.
\end{itemize}

\subsection{Metrics} Most of the cited publications have reported their results using one the following metrics: \par
For place recognition results:
\begin{itemize}
    \item Recall@1\%: represents the percentage of cases in which the
correct query sample is ranked within top 1 percentile of possible samples.
    \item Recall@1: represents the percentage of cases in which the
correct query sample is ranked first among possible samples.
    \item $F_{1}$ Max Score: measures the accuracy of the predicted samples using
    $F_1 = 2*\frac{recall*precision}{recall+precision}$
\end{itemize} \par
For map metric localization results:
\begin{itemize}
    \item Metric error: reflects the error accumulation or drift of the localization using  $E_m = \frac{\sum_{i=1}^{N}|p_i-\overline{p_i}|}{N}$ where $p_i$ and $\overline{p_i}$ are the predicted and ground truth pose.
\end{itemize}

\subsection{Place recognition}
\begin{table}[h]

\begin{centering}


  \caption{Comparison of camera based place recognition on the CVUSA and CVACT datasets.}

\begin{tabular}{|*{5}{c|}}
      \hline
       & \multicolumn{2}{c|}{CVUSA} & \multicolumn{2}{c|}{CVACT}\\
      \hline
      & {r@1\%} & {r@1} & {r@1\%} & {r@1}\\
      \hline
      \textbf{\cite{hu2018cvm}} & {91.4} & {-} & {-} & {-}\\
      \hline
      \textbf{\cite{liu2019lending}} & {93.1} & {31.7} & {-} & {-}\\
      \hline
      \textbf{\cite{zhu2021revisiting}} & {97.7} & {54.5} & {-} & {-}\\
      \hline
      \textbf{\cite{shi2020optimal}} & {99.0} & {61.4} & {95.9} & {61.0}\\
      \hline
      \textbf{\cite{guo2022soft}} & {99.7} & {95.1} & {98.1} & {85.1}\\
      \hline
      \textbf{\cite{shi2019spatial}} & {99.6} & {89.8} & {98.1} & {81.0}\\
      \hline
      \textbf{\cite{yang2021cross}} & {99.6} & {94.0} & {98.3} & {84.8}\\
      \hline
      \textbf{\cite{zhu2022transgeo}} & {99.7} & {94.0} & {98.3} & {84.9}\\
      \hline

  \end{tabular}

\end{centering}
\end{table}

In Tab. 2, we list the results of multiple camera-based place recognition methods on the CVUSA and CVACT datasets. As mentioned before, camera-based place recognition seems to go hand-in-hand with deep learning as most (if not all) the methods proposed use it in some way to try and solve this task. The use of siamese networks seems to be prevalent, which makes sense because of their ability to learn similarity details from dual inputs. However, the latest methods show that attention mechanisms and Visual Transformers can perform even better across all metrics. It does seem however that the latest improvements in accuracy are minimal, and that some new datasets maybe needed.

\begin{table}[h]

\begin{centering}

  \caption{Comparison of the F1 Max Score of LiDAR based place recognition on the KITTI dataset.}

\begin{tabular}{|*{7}{c|}}
      \hline
       & {00} & {02} &  {05} &  {06} &  {07} &  {08}\\
      \hline
      {RINet \cite{li2022rinet}} & {0.99} & {0.94} &  {0.95} &  {1.00} &  {0.99} &  {0.95}\\
      \hline
      {Locus \cite{vidanapathirana2021locus}} & {0.95} & {0.74} &  {0.96} &  {0.94} &  {0.92} &  {0.90}\\
      \hline
      {Locnet \cite{yin2018locnet}} & {0.71} & {-} &  {-} &  {-} &  {-} &  {-}\\
      \hline
      {Locnet-r \cite{yin20193d}} & {0.99} & {0.99} & {-} &  {-} &  {-} &  {0.99}\\
      \hline
      {ScanContext \cite{kim2018scan}} & {0.75} & {0.78} &  {0.89} &  {0.96} &  {0.66} &  {0.60}\\
      \hline
      {PointNetVLAD \cite{uy2018pointnetvlad}} & {0.77} & {0.72} &  {0.54} &  {0.85} &  {0.63} &  {0.03}\\
      \hline
      {SemGraph \cite{kong2020semantic}} & {0.82} & {0.75} &  {0.75} &  {0.65} &  {0.86} &  {0.75}\\
      \hline
      {SSC \cite{li2021ssc}} & {0.95} & {0.89} &  {0.95} &  {0.98} &  {0.87} &  {0.94}\\
      \hline
      
  \end{tabular}
  
\end{centering}
\end{table}

\begin{table}

\begin{centering}

  \caption{Comparison of the F1 Max Score of LiDAR based place recognition on the KITTI-360 dataset.}

\begin{tabular}{|*{8}{c|}}
      \hline
       & {0000} & {0002} &  {0004} &  {0005} &  {0006} &  {0009}\\
      \hline
      {RINet \cite{li2022rinet}} & {0.95} & {0.99} &  {0.99} &  {0.99} &  {0.99} &  {0.99}\\
      \hline
      {Locus \cite{vidanapathirana2021locus}} & {0.90} & {0.87} &  {0.89} &  {0.85} &  {0.87} &  {0.96}\\
      \hline
      {ScanContext \cite{kim2018scan}} & {0.83} & {0.77} &  {0.81} &  {0.84} &  {0.83} &  {0.85}\\
      \hline
      {PointNetVLAD \cite{uy2018pointnetvlad}} & {0.35} & {0.34} &  {0.32} &  {0.28} &  {0.29} &  {0.33}\\
      \hline
      {SemGraph \cite{kong2020semantic}} & {0.81} & {0.78} &  {0.79} &  {0.79} &  {0.83} &  {0.84}\\
      \hline
      {SSC \cite{li2021ssc}} & {0.92} & {0.97} &  {0.97} &  {0.97} &  {0.97} &  {0.97}\\
      \hline

  \end{tabular}
  
\end{centering}
\end{table}

Tab. 3 and Tab. 4 present the results of multiple LiDAR based place recognition methods on the KITTI and KITTI-360 datasets respectively that were reported by the cited publications. For LiDAR based methods, both classical and deep learning-based methods present their own unique advantages and disadvantages: methods such as \cite{kim2018scan} offer an easy, flexible and fast method to solve this task. However, the accuracy is not always as good as we would like it to be, and the method can suffer from needing wildly different matching thresholds from one scene to the other. On the other hand, methods such as \cite{fischer2022stickylocalization} or \cite{yin20193d} offer great accuracy, but at the cost of a higher level of complexity. It seems however that the greatest methods nowadays are the semi-handcrafted, such as \cite{li2022rinet} or \cite{li2021ssc} where the best of both worlds are combined to provide the best results.

\subsection{Metric Map Localization}

\begin{table*}[t]
  \begin{center}
    \caption{Comparison of Map Metric Localization methods on the KITTI dataset.}
    \label{tab:table1}
    \resizebox{\linewidth}{!}{%
    \begin{tabular}{|c|c|c|c|c|c|c|c|c|c|c|c|c|c|c|}
      \hline
       {Method}  & {Map} & {Sensor} & \multicolumn{11}{|c|}{Sequences}\\
      \hline
      \multicolumn{3}{|c|}{} & {00} & {01} & {02} & {03} & {04} & {05} & {06} & {07} & {08} & {09} & {10}\\
      \hline
      \cite{brubaker2015map} & OSM & Monocular Camera & 16 & 893 & 8.1 & 19 & - & 5.6 & - &  15 & 45 & 5.4 & 534\\
      \hline
      \cite{brubaker2015map} & OSM & Stereo Camera & 2.1 & 5.1 & 4.1 & 4.8 & - &  2.6 & - &  1.8 & 6.0 & 4.2 & 3.9\\
      \hline
      \cite{yan2019global} & OSM & LiDAR & 20 & - & - & - & - & 25 & - & 25 & - & 25 & 180\\
      \hline
      \cite{s22145206} & OSM & LiDAR & 1.37 & - & 3.37 & - & - & 1.45 & - & 1.62 & 3.60 & 2.88 & 1.56\\
      \hline
      \cite{kim2017satellite} & Satellite Map & Monocular Camera & 4.65 & - & - & - & - & - & - & - & - & 7.69 & -\\
      \hline
      \cite{miller2021any} & Satellite Map & LiDAR & 2.0 & - & 9.1 & - & - & - & - & - & - & 7.2 & -\\
      \hline
      \cite{cattaneo2019cmrnet} & Point Cloud & Monocular Camera & 0.33 &  - & - & - & - &  - & - &  - & - &  - &  -\\
      \hline
      \cite{cattaneo2020cmrnet++} & Point Cloud & Monocular Camera & 0.21 &  - & - & - & - &  - & - &  - & - &  - &  -\\
      \hline
      \cite{kim2018stereo} & Point Cloud & Stereo Camera & 0.13 & - & 0.22 & 0.23 & 0.44 & 0.14 & 0.37 & 0.13 &  0.14 & 0.17 & 0.23\\
      \hline
      \cite{zuo2020multimodal} & Point Cloud & Stereo Camera & 0.50 & 5.70 & 0.43 & 0.54 & 0.36 & 0.33 & 0.52 & 0.19 &  2.95 & 0.21 & 0.18\\
      \hline
      \cite{chen2021overlapnet} & Point Cloud & LiDAR & 0.81 & 0.88 & - & - & - & - & - & - & - & - & -\\
      \hline

   \end{tabular}
    }
  \end{center}
\end{table*}

We present the results of multiple metric map localization methods on the KITTI dataset in Tab. 5, using different visual sensors and maps from various modalities. While LiDAR-on-LiDAR localization seems to be the most popular and accurate approach in the autonomous driving industry today, stereo vision localization on point cloud maps seems to have a lot of potential. With a sub-0.5m error on most KITTI sequences using two different methods, it becomes impossible to ignore this sensor combination. While 3D maps can still be constructed by car manufacturers and their associates using LiDARs to guarantee maximum accuracy and density, it is possible to image a scenario where the cars that are meant for the consumers only feature stereo cameras and no LiDARs, bringing down the overall cost of the vehicle and data processing time tremendously, while still achieving accurate vehicle localization. It is clear however, that point cloud maps still deliver the best results, regardless of the sensor that was used on the vehicle. Other cross-modal configuration can still be useful sometimes when only a single type of map is available and can serve as an initial guess to a more advanced SLAM system.

\section{Conclusion}
In this paper we listed, compared and discussed the latest advances and findings in the area of visual map localization. We divided the visual map localization task into two major steps: place recognition and metric map localization. We explored LiDAR, camera and cross-modal based methods using multiple datasets. \par
We found that cameras can be very effective and accurate in solving the place recognition task, using deep learning mainly, making it possible to find the initial position of a vehicle in a pre-built map much more efficiently. For the metric map localization stage, point cloud maps are still essential in order to produce the most accurate results, regardless of which sensor was equipped onto the vehicle. However, the cross-modal method using stereo camera sensors and LiDAR point cloud maps seems to produce the most promising results in terms of metric map localization performance. In addition, this combination can lead to a drastic cost reduction in production and increase the accessibility of such vehicles to the general public by making it easier and cheaper to produce smart vehicles capable of accurately localizing themselves in pre-built visual maps.

{
\bibliographystyle{ieee_fullname}
\bibliography{egbib}
}

\begin{IEEEbiography}[{\includegraphics[width=1in,height=1.25in,clip,keepaspectratio]{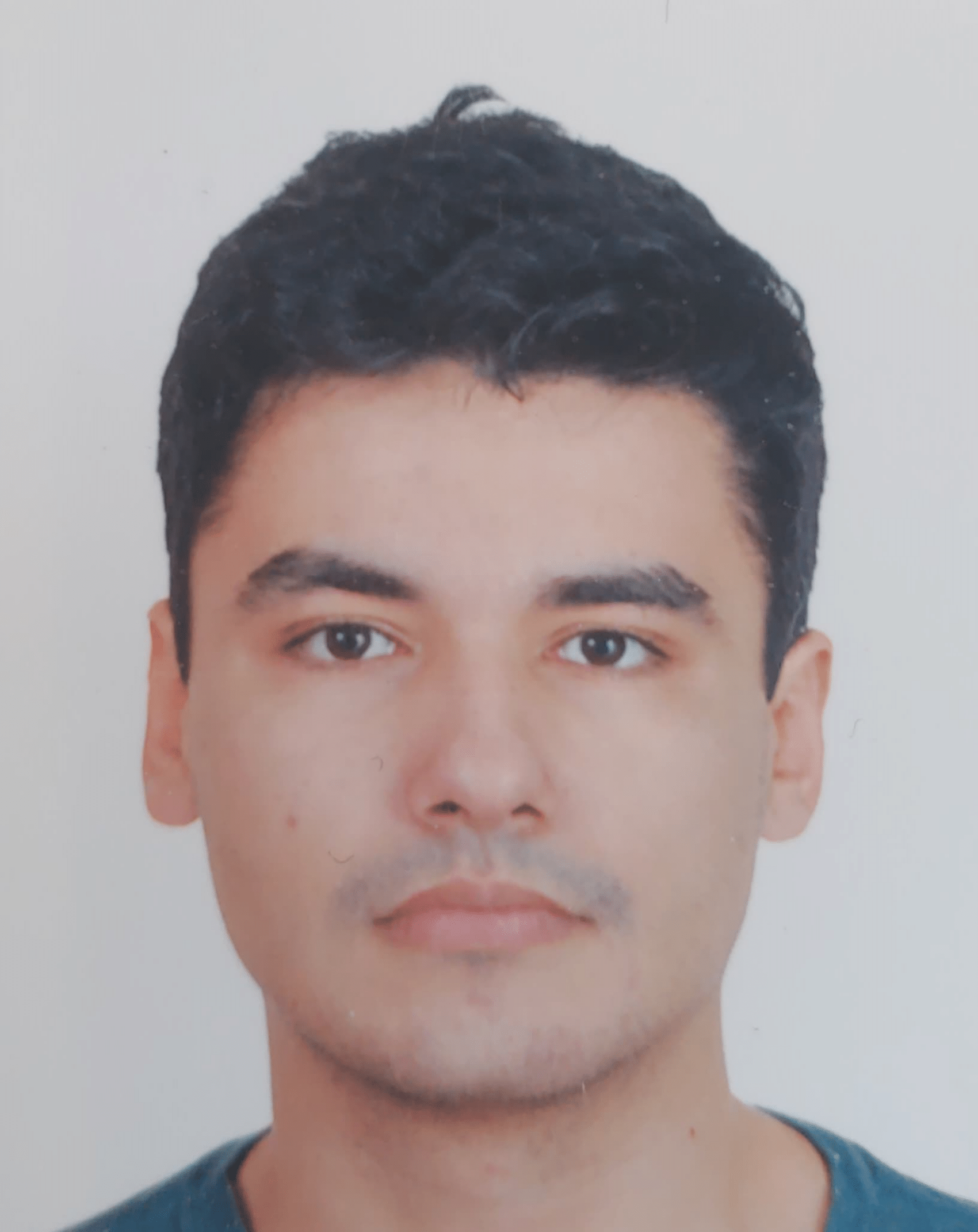}}]{Mahdi Elhousni}

is currently pursuing a PhD in Electrical and Computer Engineering at the Worcester Polytechnic in Worcester, MA, USA. Before joining WPI, he had received a BS in computer science and a MS in embedded systems from the National school For Computer Science in Rabat, Morocco. His main research interest are computer vision, deep learning and SLAM. \par
  
\end{IEEEbiography}

\begin{IEEEbiography}[{\includegraphics[width=1in,height=1.25in,clip,keepaspectratio]{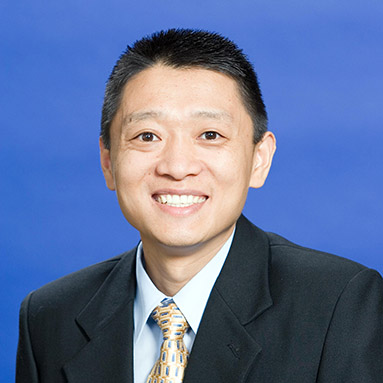}}]{Xinming Huang} 

received his Ph.D. degree in electrical engineering from Virginia Tech, in 2001. He was a Member of Technical Staffs with the Wireless Advanced Technology Laboratory, Bell Labs of Lucent Technologies. Since 2006, he has been a Faculty Member with the Department of Electrical and Computer Engineering, Worcester Polytechnic Institute (WPI), where he is currently a Full Professor. His main research interests include the areas of circuits and systems, with an emphasis on reconfigurable computing, wireless communications, information security, computer vision, and machine learning.\par
  
\end{IEEEbiography}

\end{document}